\documentclass[a4paper, 10pt, conference]{ieeeconf}      
\IEEEoverridecommandlockouts                              
\usepackage[dvipdfmx]{graphicx} 
\usepackage{url}

\title{\LARGE \bf
Hospitable Travel Agent Dialogue Robot: \\
Team Irisapu Project Description for DRC2022
}

\author{Kazuya Tsubokura$^{1}$ Fumiya Kishi$^{1}$, Kotomi Narita$^{2}$, Takuya Takeda$^{2}$  and Yurie Iribe$^{2}$
\thanks{$^{1}$ Graduate School of Information Science and Technology, 
	Aichi Prefectural University, 1522-3 Ibaragabasama, Nagakute-shi, Aichi, 480-1198, Japan
	{\tt\small \{im212008, im222005\}@cis.aichi-pu.ac.jp}}
\thanks{$^{2}$ School of Information Science and Technology, 
	Aichi Prefectural University, 1522-3 Ibaragabasama, Nagakute-shi, Aichi, 480-1198, Japan
	{\tt\small \{is201099, is191056\}@cis.aichi-pu.ac.jp, iribe@ist.aichi-pu.ac.jp}}
}

\begin{document}

\maketitle
\thispagestyle{empty}
\pagestyle{empty}

\begin{abstract}
This paper describes the dialog robot system designed by Team Irisapu for the preliminary round of the Dialogue Robot Competition 2022 (DRC2022). 
Our objective was to design a hospitable travel agent robot. 
The system we developed was ranked 8th out of 13 systems in the preliminary round of the competition, but our robot received high marks for its naturalness and likeability. 
Our next challenge is to create a system that can provide more useful information to users.
\end{abstract}

\section{INTRODUCTION}

This paper describes the travel agent dialog robot system designed by Team Irisapu for the preliminary round of the Dialogue Robot Competition 2022 (DRC2022),  an international competition evaluating the performance of dialogue robots \cite{drc1, drc2}. 
The objective of this year's competition was to create the best travel agency counter sales dialogue robot.

Our team attempted to develop a robot that would assist travel customers with an attitude of hospitality. 
Specifically, we implemented a robot system based on the hospitality guidelines of a Japanese non-profit organization that provides etiquette advice, including recommended customer service behavior \cite{JSMA}. 
The major innovations of our robot are adjusting the dialogue strategy based on the user's age, and utilizing multi-modality elements such as friendly facial expressions and polite gestures.

We describe the design policy and innovations of our system in Section \ref{sec2} of this paper. 
The flow of the dialogue scenario is explained in Section \ref{sec3}. 
The results of the evaluations provided by users are reported and discussed in Section \ref{sec4}, and the findings of this paper are summarized in Section \ref{sec5}.

\section{Design Policy and Innovations} \label{sec2}

Hospitality is one of the most important elements of customer service. Therefore, we thought it necessary to emphasize an attitude of hospitality in the travel agency customer service android robot we developed for the DRC2022 competition. We attempted to include the following six points suggested by the Japan Service Manner Association as representations of hospitality \cite{JSMA}.

\begin{enumerate}
\item{}Kind facial expressions and tones of voice 
\item{}Polite gestures 
\item{}Considerate comments 
\item{}Polite and positive language 
\item{}Ask the customer for feedback 
\item{}Pay close attention to the customer's situation 
\end{enumerate}

The first two attributes (kind facial expressions/tones of voice and polite gestures) were achieved by controlling the android robot's facial expressions and the prosody of its synthetic voice, and through gesture generation, as described in Section \ref{multimodality}. 

Attributes three and four (considerate comments and  polite/positive language) were achieved by pre-programming the android robot's dialogue, as detailed in Section \ref{sec3}. 
Furthermore, as described in Section \ref{age_strategy}, we designed the android robot to vary the way it interacted with users based on their age, in order to create a positive impression. 

As for attribute five, the robot was programmed elicit feedback from users using the following phrases, as described in Section \ref{sec3}:

Example 1: Is it correct that the tourist attractions you are considering are A and B?

Example 2: Let me briefly introduce two tourist attractions to you. 

Attribute 6, paying close attention to the customer's situation, was achieved by having the robot give back-channel feedback, as described in Section \ref{multimodality}, and by expressing empathy or by parroting what the user said, as described in Section \ref{sec3}, in order to demonstrate that the robot was listening to what the user was saying.

\subsection{Dialogue strategies based on the age of the user} \label{age_strategy}

Our system divided users into three age groups: pre-teens and teens, 20 to 39 year-olds, and those 40 or older. 
Since the android robots used in this competition are generally considered to be in their 20s to 30s, when we divided users into three age groups we made the first group younger than the robot, the second group the same age as the robot, and the third group older than the robot. 
We then adjusted the degree of politeness of the robot's utterances to correspond to the age difference between the user and the robot. 
For example, teenagers in Japan do not often encounter the use of honorific (respectful) Japanese, so we thought that if the robot used honorific expressions too often with these users it would create a barrier between the user and the robot. 
In addition, since adolescents are prone to shyness, we considered it necessary to ease their anxiety by speaking to them as informally as possible. 
Therefore, when the user was a teenager, we intentionally reduced the politeness of the robot's utterances so that the atmosphere of the dialogue would not become too stiff. The robot addressed the user with the suffix ``san'' instead of ``sama'', for example, and used normal speech level.
Note that speech level is the style in the Japanese spoken language, which appears in the main clause predicate \cite{beuchmann}.

Similarly, we lowered the politeness level of the robot's speech slightly when addressing people in their 20s and 30s, because we felt that if its speech were too formal it would come across as condescending, which could lower the user's level of satisfaction with the dialogue. 
However, when the user was 40 or older, we increased the politeness level of the robot's speech, since the user was older than the apparent age of the android robot, and thus accustomed to being treated with greater respect, especially by salespeople. 
Furthermore, the robot slowed its speech rate when the user was older than 60. 
We also lowered the pitch of the robot's voice when interacting with these users, to make its synthesized speech easier to understand.

\subsection{Utilization of multimodality} \label{multimodality}

For the purposes of this competition, the android robot is expected to interact naturally with the user. 
Since a human-like android robot is being used, users will expect it to behave like a human. 
For this reason, we utilized the following multimodal elements that the competition allowed us to control; the robot's gestures, eye gaze, facial expressions, and prosody (speech synthesis parameters).

\subsubsection{Gestures and Gaze}

\begin{figure}[thpb]
  \centering
  \includegraphics[scale=0.4]{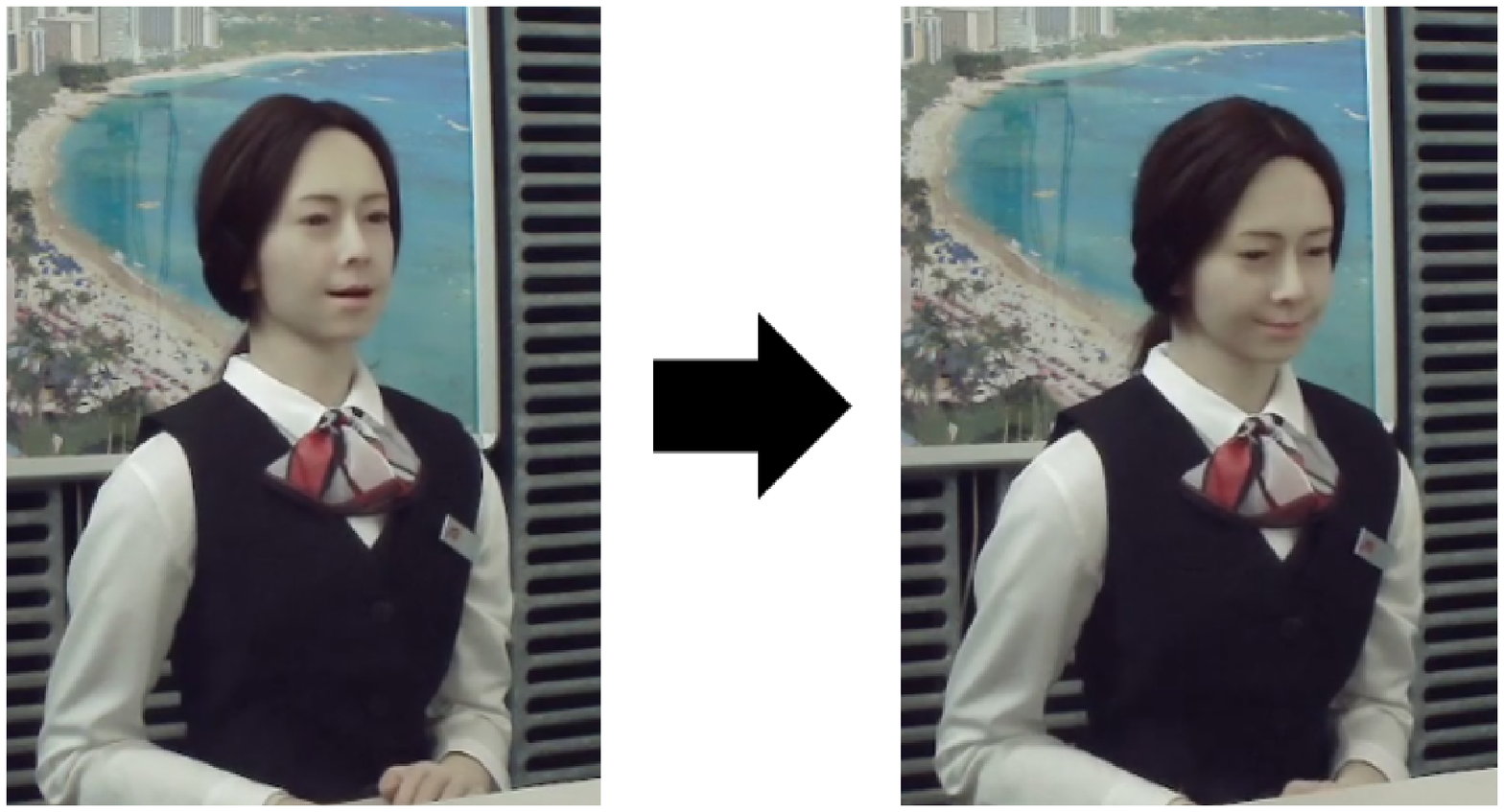}
  \caption{Bowing}
  \label{fig:ojigi}
\end{figure}

Bowing is one of the most important gestures in Japanese customer service, so the robot bows during greetings and acknowledgments. The motion used for bows was provided by the competition organizer and preloaded into the system (Fig. \ref{fig:ojigi}).  
In addition, when the user was speaking, the robot was made to appear to listen by having it nod at random times while leaning forward slightly.

\begin{figure*}[thpb]
  \centering
  \includegraphics[scale=0.5]{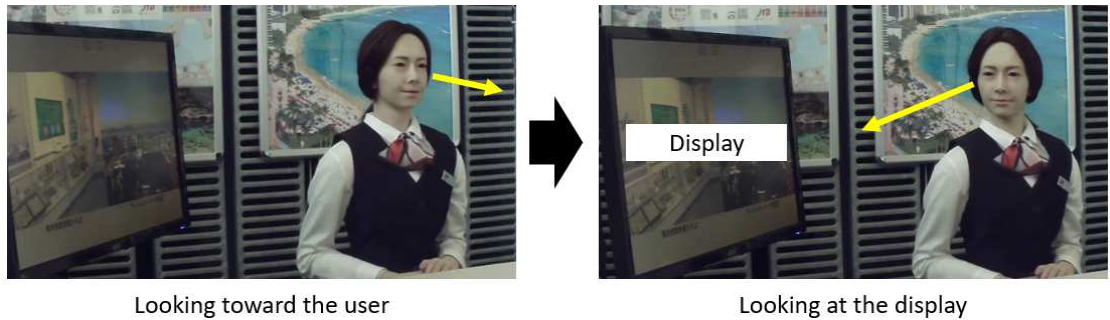}
  \caption{Looking at the display}
  \label{fig:display}
\end{figure*}

Pictures of the two sightseeing spots the user selected were displayed on the left side of the screen. 
When the robot was describing the sightseeing spots, the users were encouraged to look at the display by having the robot turn its body towards the display and gazing in that general direction (Fig. \ref{fig:display}).

\subsubsection{Facial Expression}

\begin{figure}[thpb]
  \centering
  \includegraphics[scale=0.5]{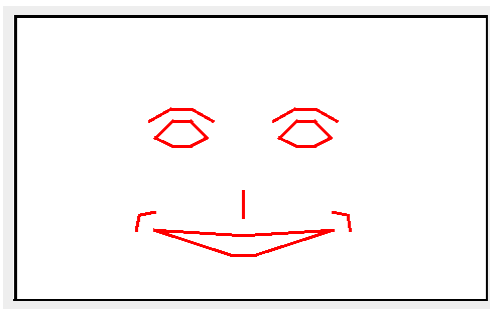}
  \caption{Normal smile}
  \label{fig:smile1}
\end{figure}

\begin{figure}[thpb]
  \centering
  \includegraphics[scale=0.5]{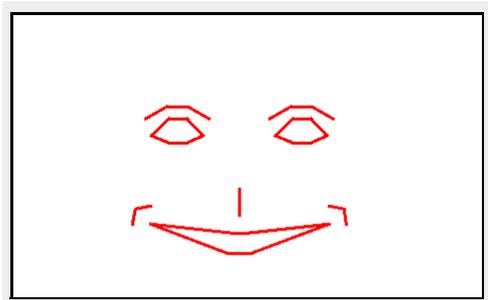}
  \caption{Broader smile}
  \label{fig:smile2}
\end{figure}

Since this competition involves a customer service task, it was desirable to have the robot smile at all times. However, having the robot maintain the same smile constantly reduces the richness of its expression and may bore the users. Therefore, we prepared two smile patterns; a normal smile (Fig. \ref{fig:smile1}), and a broader smile (Fig. \ref{fig:smile2}).

\subsubsection{prosody}

As described in Section \ref{age_strategy}, we adjusted the speed and pitch of the robot's synthesized voice according to the user's age. 
In addition, the robot raised the pitch of its voice when empathizing or expressing happiness. 
When recommending sightseeing spots, the pitch of the robot's voice was raised when introducing a recommended sightseeing spot and lowered when introducing a non-recommended sightseeing spot, thus implicitly identifying the preferred recommendation.

\section{Dialogue Flow} \label{sec3}

\begin{figure*}[thpb]
  \centering
  \includegraphics[scale=0.6]{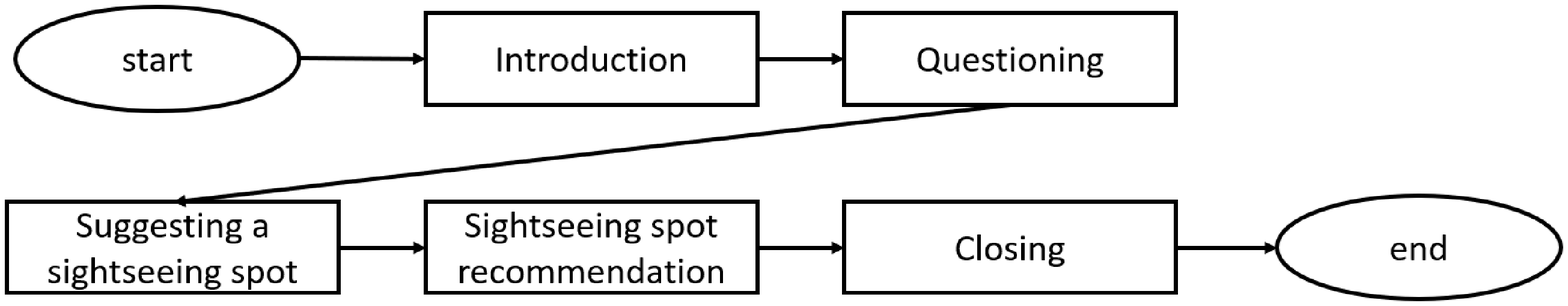}
  \caption{Dialogue flow}
  \label{fig:df}
\end{figure*}

As shown in Fig. \ref{fig:df}, our system scenario is divided into five phases, which guided the android's utterances. Questions asked by the user terminated the robot's utterances. 
The details of each phase of the scenario are described below.

\subsection{Introduction}

The introduction phase consists of a greeting, a self-introduction, and identity questions for the user. The robot first greets users with, ``Hello. Thank you for coming in the hot weather today,'' to encourage the user to speak. During the self-introduction, we intended to make the robot more familiar by having it say its name. The robot also says, ``Actually, this is my first day on the job, so I may be unfamiliar with some aspects of the job, but please be patient with me.'' This was intended to evoke a sense of inexperience, in order to reduce user discomfort in the event of possible inappropriate responses in subsequent dialogues.

Finally, the robot elicits the name of the user. The robot can repeat the user's name during the subsequent dialogue if the user's name is correctly elicited. We expected the user to respond to this question by saying, ``I am X Y,'' ``My name is X Y,'' or by giving only his or her name. We conducted morphological analysis of the users' replies using GiNZA \cite{ginza}, an open-source library for natural language processing of Japanese, then used the words tagged ``noun-person-name'' as the user's name. If there was no word assigned the tag ``noun-proper noun-person's name'' after morphological analysis, we had the robot address the user as ``okyaku-sama (customer)'', which is a common, respectful form of address in Japanese. 

\subsection{Questioning}

In the questioning phase, we attempted to obtain information on user preferences in order to interact with visitors in an enjoyable way, and to recommend appropriate sightseeing spots. We prepared the following six questions:

\begin{enumerate}
\item{}Have you ever had a conversation with an android robot? 
\item{}Who are you traveling with? 
\item{}Are you an indoor or an outdoor person? 
\item{}Do you prefer Japanese or western food?
\item{}Do you like taking pictures? 
\item{}Are you interested in history? 
\end{enumerate}

Asking these six questions one after another would appear mechanical to the user, so we sought answers to the six questions in conjunction with parroting, sympathizing, and synchronized responses. 

Specifically:

Robot: Are you an indoor or an outdoor person? 

User: I'm an indoor person. 

Robot: You're an indoor person? Me too! 

And:

Robot: Do you prefer Japanese or western food? 

User: I prefer Japanese food. 

Robot: Japanese food is good. 

\subsection{Suggesting a sightseeing spot}

In the phase when sightseeing spots are suggested, the robot describes two local sightseeing spots, and users are asked which of the two locations they are interested in. The descriptions used when suggesting these sightseeing spots were taken from a sightseeing spot database provided by the DRC2022 competition organizers. The provided descriptions were too long to be read aloud in their entirety however, so they were summarized using OpenAI's API \cite{openai}. They were also converted from regular to honorific forms using simple rules (e.g. ``shite-iru'' $->$ ``shite-imasu'').

\subsection{Sightseeing spot recommendation}

The robot describes two recommended sightseeing spots in the sightseeing spot recommendation phase, based on the user's responses to the questions, ``Do you like taking pictures?'' and ``Are you interested in history?'' Although the robot also asked other questions during the questioning phase, to simplify selection of the recommendation text, our system does not consider the users' responses to the other questions.
If the user likes taking pictures, and one of the following types of local attractions is included in the provided sightseeing spot database, the robot will say, ``Since you like taking pictures, $<$recommended tourist attraction$>$ is recommended.''

\begin{itemize}
\item{}Gardens/Botanical Gardens/Herbarium 
\item{}Parks 
\item{}Towers/Observation facilities 
\item{}Rivers/Waterfalls/Springs/Valleys 
\end{itemize}

If the user is interested in history and one of the following types of local attractions is included in the provided sightseeing spot database, the robot will say, ``Since you are interested in history, we recommend $<$recommended sightseeing spot$>$.''
\begin{itemize}
\item{}Temples/Shrines/Churches 
\item{}Art museums/Galleries 
\item{}History museums/Science Museums/Archives 
\item{}Factory/Facility tours 
\end{itemize}

If the above conditions are not met, a manually prepared tourist attraction recommendation sentence is uttered, based on the tourist attraction category of the recommended tourist attraction.

Finally, users were guided to choose the recommended sightseeing spot during a follow-up exchange, by reminding them, ``I hope you are interested in the recommended sightseeing spot.''

\subsection{Closing}

The dialogue is terminated when the dialogue flow described above is completed, or the specified dialogue time has elapsed.

\section{Result of the preliminary round} \label{sec4}

This section describes the results of evaluations conducted by users during the preliminary DRC2022 competition, after using the systems developed by the competitors. Their evaluations were based on the effectiveness of the android's recommendations and the satisfaction of the users. The effectiveness of the android's recommendations was measured by having users assign a value between -100 and 100 to reflect how much the user's desire to visit the recommended sightseeing spot increased after their dialogue with a particular robot. The following nine items were evaluated on a 7-point Likert scale, where 1 represented strong disapproval and 7 represented strong approval:

\begin{enumerate}
\item{}Satisfaction with choice: ``Were you satisfied with your choice of tourist attraction to visit?''
\item{}Informativeness: ``Were you able to obtain sufficient information about the sightseeing spots?''
\item{}Naturalness: ``Did you have a natural dialogue with the robot?''
\item{}Appropriateness: ``Was the robot's service appropriate?''
\item{}Likeability: ``Was the robot's service likable?''
\item{}Satisfaction with dialogue: ``Were you satisfied with your interaction with the robot?''
\item{}Trustworthiness: ``Did you trust the robot?''
\item{}Usefulness: ``Did you use the information obtained from the robot to select the sightseeing spot?''
\item{}Intention to reuse: ``Would you like to visit this travel agency again?''
\end{enumerate}

The results of the impression evaluation questionnaire are shown in Table \ref{tab:result}, where ``Avg.'' represents the averaged values of the users' evaluation responses for the Team Irisapu android robot, and ``Ranking'' indicates the attribute rankings for the Team Irisapu robot among the 13 robots (including the baseline robot) which competed in the preliminary round of the competition. The total impression rating for our robot was 39.76, which means our robot was ranked 8th among the 13 robots.

For questionnaire items \#3 (Naturalness) and \#5 (Likeability), our robot received the higher evaluation among the 13 robots. We think our age-appropriate dialogue strategy and the multimodality responses described in this paper contributed to the robot's naturalness and likeability scores.

On the other hand, for questionnaire item \#8 (Usefulness) our robot was ranked 12th, suggesting that our system's utterances did not supply enough beneficial information to users. This may be because our system was not able to provide sufficient information when selecting appropriate sightseeing spots, because not enough of the users' responses were utilized when selecting sightseeing recommendations. As a result, our robot's performance during the sightseeing spot recommendation phase was insufficient. The average Trustworthiness (item \#7) of our android's recommendations was 11.29 (5th out of 13 robots), suggesting that our system provided more accurate recommendations than some of the other teams' systems.

\begin{table}[h]
\caption{Results of the Questionnaire}
\label{tab:result}
\begin{center}
\begin{tabular}{c|ccccccccc} \hline
Item & Avg. & rank \\\hline \hline
Satisfaction with choice & 4.36 & 8 \\\hline
Information & 4.36 & 8 \\\hline
Naturalness & 4.16 & 4 \\\hline
Appropriateness & 4.36 & 7 \\\hline
Likeability & 5.08 & 3 \\\hline
Satisfaction with Dialogue & 4.64 & 6 \\\hline
Trustworthiness & 4.56 & 5 \\\hline
Usefulness & 4.24 & 12 \\\hline
Intention to reuse & 4.00 & 10 \\\hline
\end{tabular}
\end{center}
\end{table}

\section{CONCLUSIONS} \label{sec5}

The results of the user impression questionnaire used to evaluate the travel agent dialog robot systems submitted for the DRC2022 competition ranked Team Irisapu's android robot 8th overall among 13 robots (including the baseline robot) evaluated during the preliminary round of the competition. However, our robot was ranked 3rd in likeability and 4th in naturalness. Our robot's weak point was the usefulness of its sightseeing recommendations to users, which was likely due to not utilizing enough customer feedback when recommending sightseeing spots. In the future, it will be necessary to improve the usefulness of the information the robot provides, as well as the quality of its dialogue, in order to improve our robot's ratings.


\end{document}